# D2E: An Autonomous Decision-Making Dataset involving Driver States and Human Evaluation

Zehong Ke[#], Yanbo Jiang[#], Yuning Wang, Hao Cheng, Jinhao Li, and Jianqiang Wang*

*Abstract*— With the advancement of deep learning technology, data-driven methods are increasingly used in the decision-making of autonomous driving, and the quality of datasets greatly influenced the model performance. Although current datasets have made significant progress in the collection of vehicle and environment data, emphasis on human-end data including the driver states and human evaluation is not sufficient. In addition, existing datasets consist mostly of simple scenarios such as car following, resulting in low interaction levels. In this paper, we introduce the Driver to Evaluation dataset (D2E), an autonomous decision-making dataset that contains data on driver states, vehicle states, environmental situations, and evaluation scores from human reviewers, covering a comprehensive process of vehicle decision-making. Apart from regular agents and surrounding environment information, we not only collect driver factor data including first-person view videos, physiological signals, and eye attention, but also provide subjective rating scores from 40 human volunteers. The dataset is mixed of driving simulator scenes and real-road ones. High-interaction situations are designed and filtered to ensure behavior diversity. Through data organization, analysis, and preprocessing, D2E contains over 1100 segments of interactive driving case data covering from human driver factor to evaluation results, supporting the development of data-driven decision-making related algorithms.

## I. INTRODUCTION

Autonomous driving technology has been considered to bring a significant revolution in the future of transportation. Typically, autonomous driving technology can be divided into perception, decision-making, and control. Among them, decision-making under complicated traffic scenarios is a critical challenge towards high-level automated driving [1, 2]. Existing decision-making methods can be categorized into rule-driven and data-driving approaches. While rule-based methods are reliable and easily implementable in low-level autonomous driving scenarios, they lack generalizability in complex scenarios [3]. Data-driven approaches, such as deep learning and reinforcement learning have garnered widespread attention due to the potential for generalization [4].

The effectiveness of data-driven methods highly relies on the quality of the datasets used. While the volume and variety of datasets have explosively increased in recent years to accelerate decision-making technology, there remains a relative scarcity of human-factor data [5, 6]. Even when some datasets include human-factor data such as gaze points [7], it is often isolated from the driving environment. Human factor data with driving environment information contributes to a better understanding of human driving intelligence, facilitating the development of autonomous driving decision-making.

On the other hand, current datasets are increasingly focused on challenging scenarios, which are difficult to be satisfied through real-world vehicle collection. The challenging scenarios in real-world driving are difficult to replicate and guarantee safety. In comparison, simulator-based data collection offers higher safety and efficiency, allowing for scenario repetition while it may lack authenticity. Therefore, a more comprehensive dataset should include high-risk events from both simulator and real-world driving to mutually validate and ensure data quality.

To address the above issues, we establish D2E, an autonomous decision-making dataset involving driver states and evaluation scores from human reviewers. Following the current trend of challenging scenarios [8, 9], it prioritizes high-interaction scenarios, encompassing cases in the simulator and events extracted from real-world driving segments. Specifically, the D2E dataset, as its name suggests, integrates human factor data from drivers' factors, and driving environment data, to human evaluations. In addition to recording fundamental driving environment data such as the position and speed of the ego vehicle and surrounding vehicles, D2E also collects physiological and eye-tracking information of drivers in each driving event, together with the first-person view videos recorded. Furthermore, since human drivers also make mistakes, subjective ratings given by 40 human volunteers are provided for each driving event so that the algorithms can recognize the driving performance level of the cases.

Compared with previous autonomous driving datasets, the contributions of the D2E dataset are as follows:

1) Human-factor data: Our dataset collects human-factor data for each event, including physiological data such as heart rate, blood oxygen, and eye-tracking data. Additionally, qualified raters are recruited to provide subjective scores for each driving event based on first-person-view videos.

2) Integration of simulator and real vehicle: D2E collects driving events from both simulator and real vehicle driving, including simulated data from high-risk and complex scenarios and natural driving data, ensuring enough interaction intensity while enhancing data authenticity.

3) High-interaction scenarios design: We designed 12 driving scenarios in the simulator which are all complicated

[#]These authors contributed equally
*This paper is supported by the National Natural Science Foundation of China with Awards 52131201, Beijing science and technology star program (20230484418), and Tsinghua University Toyota Joint Research Center for AI Technology of Automated Vehicles (Grant TTRS 2024-06).
Yanbo Jiang, Yuning Wang, and Jianqiang Wang are with the School of Vehicle and Mobility, Tsinghua University, 100084, Beijing, China
Zehong Ke, and Jinhao Li are with Xingjian College, Tsinghua University, 100084, Beijing, China. (e-mail: jyb23@mails.tsinghua.edu.cn; wjqlws@tsinghua.edu.cn).

and interactive events. As for the real-road driving, interactive cases are filtered manually.

The subsequent sections of the article are outlined as follows: Section II provides a review of previous literature on datasets. Section III introduces our methodology, including details on experimental equipment and data collection procedures. In Section IV, a comprehensive analysis of D2E is presented. Section V provides a brief conclusion for this article.

## II. RELATED WORKS

The traditional decision-making methods in autonomous driving were often based on fixed rules, making it difficult to adapt to complex scenarios and handle vehicle interactions, so it was difficult to achieve driver satisfaction [3]. With the development of artificial intelligence, data-driven approaches, such as deep learning and reinforcement learning, have gained increasing interest among researchers. A good dataset often determines the training results and final accuracy of neural networks [10].

From the view of traffic systems, data sources for decision-making datasets can be mainly divided into three categories [5]. The first category is vehicle information, which includes the location information of vehicles, and internal data on CAN buses such as throttle position, steering wheel angle, engine states, etc. The second category is environment data, encompassing data about other traffic participants, traffic lights, and other traffic-related details which are collected by perception sensors. The third category is driver state, including the behavior interactions between the driver and the cabin, physiological states, gaze allocation and distraction, etc.

The most common data types in existing datasets are vehicle and the environmental information. The DDD17 dataset [11] incorporates information obtained from the vehicle's onboard port, such as speed, GPS data, engine speed, transmission input torque, and fuel consumption. The EU Long-term dataset [12] integrates eleven heterogeneous sensors, including various cameras, lidar, radar, IMU, and GPS-RTK, which enables the vehicle to perceive the surrounding environment and simultaneously locate itself. In addition, there are some classic datasets available. The NGSIM dataset [13] records structured road intersections and highway entrance and exit ramps. The HighD dataset [14] captures German highway scenarios using drones. The SHPR2 dataset [15] documents over 1,900 light-vehicle crashes. The KITTI and Waymo [8, 16] datasets collect environmental data using LiDAR and cameras. Although the data source of vehicle and environment is comprehensive, the attention on driver data is insufficient, which also provides great value and inspiration for the development of autonomous driving [17].

Some datasets specifically collect driver data as a dataset for human driver pattern analysis. Xing et al. [18] collected physiological information such as skin, muscle, respiration, and blood oxygen levels from drivers during driving simulator experiments. The DMD dataset captures the faces, bodies, and hands of 37 drivers from 3 cameras, recording a total of 41 hours of data on driver distraction, gaze point, drowsiness, and hand-steering wheel interaction [19]. One major limitation of current driver datasets is their singular focus, lacking a comprehensive collection of various driver information such as behavior, attention, and physiological data. Secondly, these driver datasets lack integration with vehicle and environmental data, making it difficult to support the development of decision-making algorithms. Therefore, our dataset aims to gather integrated data on vehicles, environment, and drivers.

In addition to the lack of driver data, low interactivity is another challenge in current datasets [10]. In the Argoverse dataset[20], there are only about three hundred interesting trajectories that satisfy at least one preset condition, including intersections and lane changing. In highD dataset [14], 94.5% of the events are simple car-following. On the other hand, some datasets such as rounD [21], inD [22] stick to one kind of complicated traffic position including intersections and roundabouts so most of the cases are interactive. The low interactivity of current datasets is mainly reflected in two aspects: one is the small proportion of interactions and the other is too much emphasis on a single behavior (roundabouts, lane changes, etc.), which makes the data distribution biased.

Because of the ability to design high-interactive scenes, in recent years, driving simulator datasets have gradually appeared in various studies [23]. A driving simulator is a tool that replicates real driving scenarios in a virtual environment. In addition to being able to design scenarios, it also has many advantages, including the ability to design repeatable experiments, no security and legal risks, and rich data interfaces. The article by Wang et al. explores how to determine reasonable speed limits to ensure traffic safety through driving simulator studies in dynamic low-visibility environments [24]. However, driving simulators also have some disadvantages, the most important issue being their authenticity, although technological advances have brought simulators closer to real vehicles [25]. It's worth noting that many current datasets choose driving simulator data as their primary component, supplemented by a portion of real-vehicle datasets. This allows training on driving simulator datasets with abundant high-interaction scenarios and validation on real-vehicle datasets.

The evaluation of dataset cases is another issue usually neglected. Most studies treat human drivers as experts and assume their trajectories as ground truth, training models by minimizing the error between predicted and actual behavior [8]. However, human drivers also make errors, making the real trajectory not the best choice. In addition, different drivers also have various preferences, leading to the gap between the data and real application. Therefore, evaluations of the driving performance in each case should be given so that the users can have a comprehensive assessment of the real trajectories and adjust their algorithms accordingly [26]. In this research, we design a decision-making evaluation method and assign rating scores to all cases as additional data sources.

## III. METHOD

### A. Driver to Evaluation Dataset

In this work, we propose the D2E dataset, covering the full-cycle data source from driver states to evaluations. D2E aims to assist in making the intelligent decision-making of autonomous vehicles more aligned with the subjective experiences of human drivers. Due to the necessity of collecting data from dangerous and complex scenarios, the dataset primarily utilizes a driving simulator for driver safety and experimental repeatability and complements part of real-world driving data. As for the simulation part, 80 qualified drivers with various driving experiences and features are recruited to drive on 12 interactive scenes. The real-world driving data is collected from 7 drivers in a structured road section with high traffic flow and multiple interactions in Suzhou, Jiangsu province. After manual filtering, 153 20-second segments of hazardous scenarios are extracted. In addition, the dataset invites 40 volunteers to watch the driver's first-person view videos and score the driving performance, thereby achieving a subjective evaluation of the driving trajectory from a third-party perspective.

## B. Driving Simulator Data Collection

The experimental equipment includes a simulator platform and wearable devices, as shown in Fig. 1. The main structure of the simulator platform includes the driver cabin, scene monitors, motion platform, and cloud server. The motion platform is a three-degree-of-freedom simulation platform that can simulate the vertical, lateral, and pitch movements of the vehicle. Three display screens connected to the motion platform can faithfully reproduce the actual driving view in the left, center, and right regions. The main screen can provide images from the rearview mirror perspective and the rear mirror perspective. The driver cabin replicates all the real operational layouts of a real vehicle, including the steering wheel, accelerator and brake pedals, dashboard, seat, seatbelt, gear lever, etc., ensuring the realism of the driving experience. All data is collected, transmitted, and organized on the cloud server. In addition, we equipped the driver with eye-tracking glasses and wearable physiological recorders. The eye-tracking glasses can record the driver's first-person view and capture real-time gaze points. The wearable physiological recorder includes wrist sensors, finger sensors, and earlobe sensors, as shown in Fig. 2, which can record physiological data such as Electrodermal Activity (EDA), Skin Temperature (SKT), Beats Per Minute (BPM), Pulse Oxygen Saturation (SpO2), etc. Finally, the simulator data, eye-tracking data, and physiological data are aligned through absolute time.

The selection of scenarios in the simulator dataset is critical for the dataset. High-interaction and representative scenarios significantly enhance the richness of the simulator dataset and the generalization ability of the models trained on it. In the Waymo dataset [15], the standard of judging interaction is the existence of conflict regions among agents. In Argoverse dataset [15], interaction is preset to different categories, including intersections and lane changing. After referring to these previous popular datasets, we designed the following 12 types of scenarios (a) to (j) based on the simulator software. As shown in Fig. 3, the designed scenarios are listed below:

Fig. 3, each of these scenarios will be listed below.

- Car following with preceding car braking on urban roads
- Highway ramp merging
- Lane changing before the highway ramp exiting
- Overtaking on urban roads

- Overtaking on highway roads
- Car following with front car cutting out on urban roads
- Turning left at an unprotected intersection with pedestrians, cyclists, and other vehicles
- Turning right at an unprotected intersection with pedestrians, cyclists, and other vehicles
- Crossing an unprotected intersection with pedestrians, cyclists, and other vehicles
- Cruising with a truck from the adjacent side suddenly cutting in
- Long-term urban driving with other vehicles
- U-turn facing conflicting vehicles.

Fig. 3 (a)~(j) respectively represent the above scene conditions, in which the red vehicle in the scene is the self-car, the white vehicles are other vehicles, and some scenes also include agents such as pedestrians and cyclists. The red line represents the driving task given to the driver, and the black line represents the default trajectory and speed set for each agent. It is worth mentioning that in the simulator software, other agents have simple intelligence and can react based on the behavior of the surrounding environment.

The driving simulator dataset involved 80 licensed drivers. To encompass a wide range of decision-making outcomes and driving trajectories, efforts were made to select drivers from diverse age groups and with varying levels of driving experience. Additionally, information regarding the drivers' self-assessed driving styles and their involvement in accidents was collected through a questionnaire. The details are summarized in Table I.

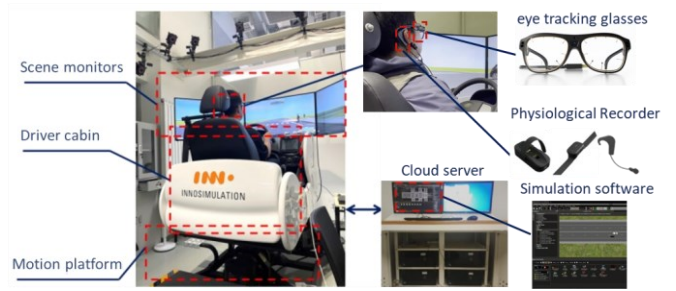

Figure 1. Composition of the wearable physiological recorder.

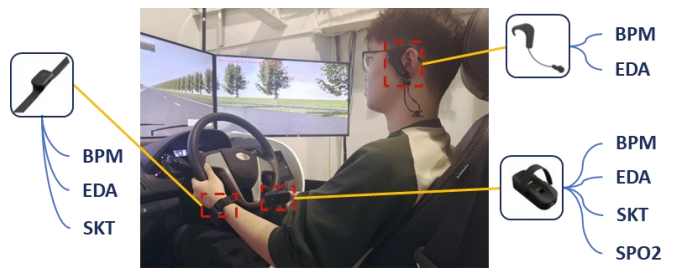

Figure 2. Composition of the wearable physiological recorder.

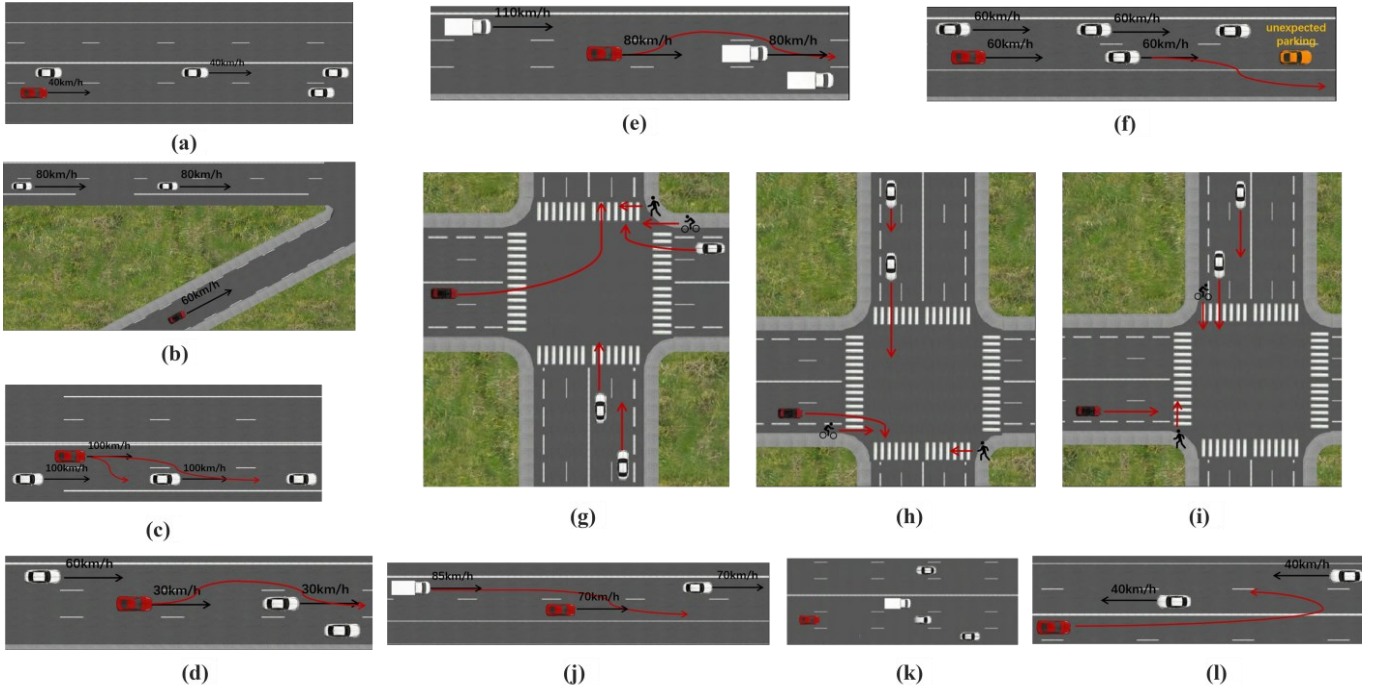

Figure 3.   Experimental equipment for driving simulator dataset.

| | Feature | Number | Ratio |
|---|---|---|---|
| Gender | Male | 68 | 85% |
| | Female | 12 | 15% |
| Age | 18-30 | 23 | 28.75% |
| | 31-40 | 27 | 33.75% |
| | 41-50 | 17 | 21.25% |
| | 50+ | 13 | 16.25% |
| Driving years | 0-3 | 15 | 18.75% |
| | 4-9 | 30 | 37.5% |
| | 10-19 | 28 | 35% |
| | 20+ | 7 | 8.75% |
| Driving style | Conservative | 35 | 43.75% |
| | Mediums | 36 | 45% |
| | Aggressive | 9 | 11.25% |
| Accident experience | None | 53 | 66.25% |
| | Minor accident (e.g., scratches) | 20 | 25% |
| | Moderate accident (e.g., rear-end collision) | 7 | 8.75% |

From the result of the basic questionnaire, the gender ratio of the drivers is close to the gender ratio of driving miles in China. The age distribution is broad and balanced, covering young people, the middle-age, and seniors. Also, the distribution of driving experience is wide. 43.75% of the participants drive for more than 10 years. As for the self-assessment of driving styles, drivers generally tend to rank themselves more conservatively. Regarding involvement in accidents, all selected drivers have no history of major accidents, and more than half of the drivers claim no involvement in any accidents.

## C. Real-vehicle Data Collection

To enhance the diversity and authenticity of the dataset, and to ensure that models trained on the driving simulator dataset can be tested on the real-vehicle dataset, we collected real-vehicle data from 7 drivers in Suzhou, China. The route is shown in Fig. 4, the traffic flow in this route is moderate, which ensures the existence of complex interactive behaviors while preventing the perception results from deteriorating due to too many traffic participants. More importantly, this section is equipped with roadside sensing equipment and V2X systems, as shown in Fig. 5. To align as closely as possible with the data types in the simulator dataset, our vehicle was equipped with IMU and GPS modules, and roadside devices such as cameras and radars will utilize perception algorithms to obtain information such as the positions, velocities, and accelerations of other vehicles in the current environment, excluding the ego vehicle, each of these vehicles is assigned a unique and continuous ID, and both the ID and information are transmitted to our vehicle in real-time via V2X devices, as shown in Fig. 6. Additionally, for consistency, each driver also wore eye-tracking glasses and physiological monitoring devices. Finally, Finally, relevant preprocessing is performed on these data to obtain the same type of data as the simulator dataset.

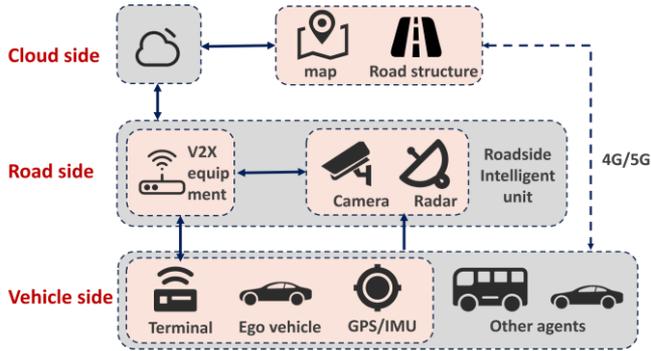

Figure 5.   Roadside sensing equipment and V2X systems.

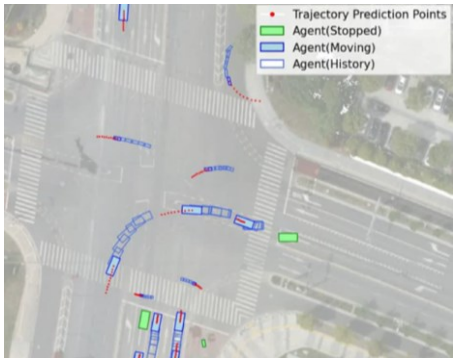

Figure 6.   Visualization of environmental vehicle information received by ego vehicle through V2X

During the collection process, we manually recorded moments of complex interactive scenes occurring during the driving process and captured 10 seconds before and after each moment, resulting in a total of 153 segments of 20-second videos. AN example of the first-person view of the real-vehicle dataset and the driving simulator dataset is shown in Fig. 7.

### D. Human Evaluation

After collecting the driving data from the driving simulator and real vehicle, we recruited 40 third-party volunteers as assessors to serve as experts for evaluation and rating. To be noted that some volunteers do not possess a driver's license because passengers also have the right to assess the vehicle's decisions, the details of assessors are summarized in Table Ⅱ. To standardize the assessors' ratings, we set scoring rules based on a percentage scale, a score of 90 indicates excellent driving behavior, while 60 represents barely acceptable driving behavior, and major accidents resulting in collisions, etc., should receive a score of 0. Additionally, to ensure relative fairness and consistency in scoring, each assessor needed to consecutively rate 80 driving trajectories of the same scenario type. To prevent potential judgment errors due to fatigue, scoring for each assessor was limited to one driving simulator

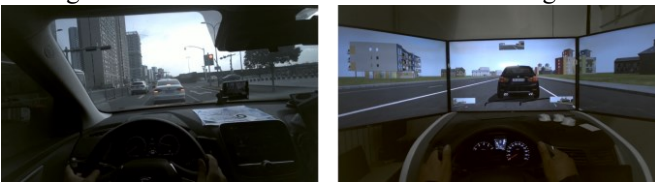

Figure 7.   The first-person view of real-vehicle dataset and of driving simulator dataset

scenario per day.

TABLE II.   Assessor distribution of driving simulator dataset

| | Feature | Number | Ratio |
|---|---|---|---|
| Gender | Male | 20 | 50% |
| | Female | 20 | 50% |
| Drving license | Have | 26 | 65% |
| | Don't have | 14 | 35% |

## IV. Dataset Organization and Analysis

This section will introduce the data organization structure, outline the data filtering rules, and conduct an analysis of evaluation scores.

### A. Data Organization Format

The tree diagram in Fig. 8(a) illustrates the data organization structure of the D2E dataset. At the root directory, the data is classified based on its platform into 'simulator' and 'real_vehicle', namely. For real vehicle data, it is further categorized based on the driver's ID (01~07). Within each folder named by driver's ID, 'assessment' is a table containing the evaluation scores for their driving events. Human factor data and perception data are organized based on driving events. The detailed organization of human factor data is illustrated in Fig. 8(b), where the naming convention for eye-tracking and physiological data is based on the collection location and signal type. For instance, 'Left Pupil' means eye-tracking data for the left eye and 'Ear9 BPM' represents the heart rate data collected from the ear. For simulator data, the data is categorized based on scenes. Within each folder named by scene number, there are both the original evaluation table and the evaluation table after data filtering for that specific scene. The organization of human factor data and driving data is similar to that of real vehicle data, but the categorization is based on the driver's ID.

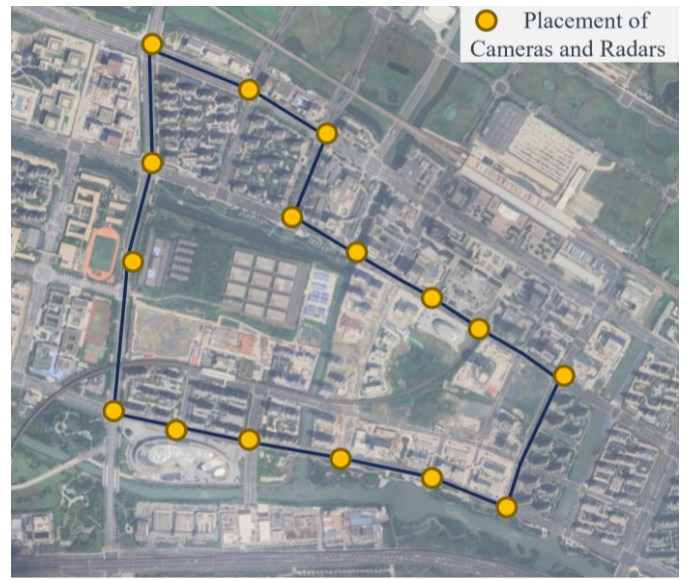

Figure 4.   The route of real-vehicle dataset and sensor placement.

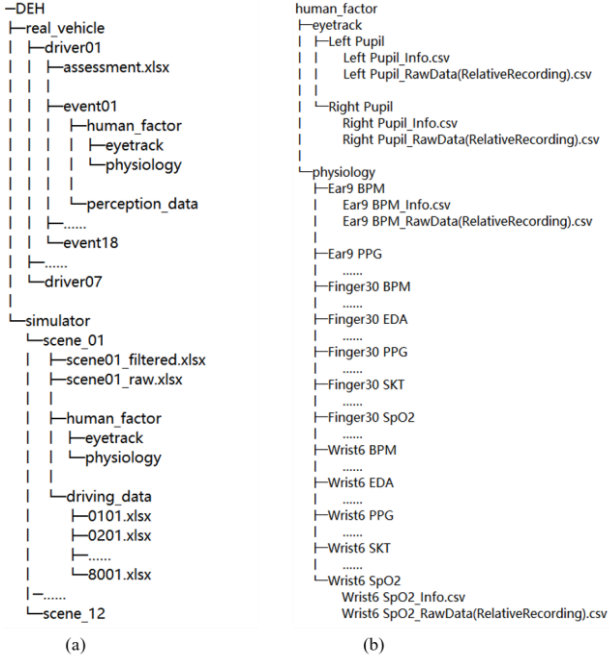

Figure 8.   The data organization structure for the DHE data (a) and human factor data (b)

## B.  Data Filtering

Data filtering involves two main components: data slicing and data screening. Data slicing is performed to extract data corresponding to the time of driving events, eliminating redundant information outside the timeframe of the driving events. Data screening aims to enhance data reliability, particularly in the case of evaluation scores, which are significantly influenced by the seriousness of assessors.

In simulator data collection, eye-tracking data is collected for each driving event while physiological data is gathered according to individual drivers, requiring slicing according to the time intervals of driving events. Due to the first column of the physiological data table representing relative time, it is necessary to calculate the relative time by subtracting the starting recording time from the absolute time of each driving event. Afterward, physiological data for each driving event can be obtained with sliced tables. In real vehicle data collection, both eye-tracking and physiological data are collected based on individual drivers, not according to driving events. Therefore, slicing needs to be performed based on the time of high-interaction events mentioned in section III, using a similar approach.

As for the evaluation data, filtering is needed since some raters may lose focus and give random scores. We implemented a two-step filtering approach: scene-specific filtering and comprehensive veto. In each scene, assessors who gave scores above 70 for events that received zero scores from over 14 other assessors more than 3 times were identified. This aimed to focus judgments on significant collision events and avoid excessive filtering to respect assessors' subjective judgments on minor collisions. After applying this filter across all scenes, a statistical analysis was conducted. If a driver received poor ratings in more than 1/3 of the scenes, their score reliability was considered low, and all scores from that driver were disregarded. If not, it was assumed that their ratings were reliable in other scenes, even if they were distrustful in a specific scene.

Finally, all the aforementioned types of data are aligned in terms of time. After data filtering, the simulator data retained records for 960 (12*80) driving events, including information on both the ego and surrounding vehicles' speed, position, ego vehicle acceleration, etc. Additionally, there are 684 (12*57) physiological data records, 960 (12*80) eye-tracking data records with first-person view videos, and 33079 valid assessment scores. The absence of physiological data for all events is due to the incorrect wearing of physiological devices by some drivers, leading to data invalidation. For real vehicle data, all 153 driving events are equipped with information regarding ego and surrounding vehicle perception, drives' physiological and eye-tracking data, as well as 40 evaluation scores, totaling 6120 scores.

## C.  Evaluation Data Analysis

Table III lists the distribution frequency of the scoring of 12 scenes on the simulator with the highest two items marked down and the lowest one displayed in bold. Intuitively, the frequency of ratings between 65 and 70 tends to be lowest, and the ratings tend to be concentrated in the interval of less than 60 or more than 90, which respectively corresponds to the failing and excellent intervals in the scoring rules. The latter proves that the driving scenarios we designed are indeed challenging, and the drivers who were able to complete them successfully are also considered to be excellent. It could be further investigated whether these distributional trends are related to the mechanism of subjective evaluation, i.e., whether evaluators are less inclined to give an ambiguous evaluation. Fig. 9 depicts the distribution of evaluation scores in both simulator and real vehicle scenarios. The distribution of scores given by assessors above 65 is similar, while the proportion of scores below 65 in the simulator significantly exceeds that in the real vehicle. This is attributed to the fact that most scenarios simulated are high-risk, with a higher probability of collision accidents, leading to a higher proportion of lower

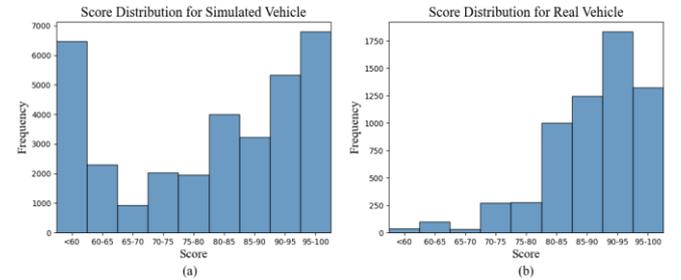

Figure 9.   The distribution of evaluation scores for simulator(a) and real vehicle (b)

TABLE III.     EVALUATION SCORES FREQUENCY FOR EACH SCENE ON THE SIMULATOR

| Scene ID | <60 | 60-65 | 65-70 | 70-75 | 75-80 | 80-85 | 85-90 | 90-100 |
|---|---|---|---|---|---|---|---|---|
| 01 | 0.363 | 0.094 | **0.030** | 0.066 | 0.049 | 0.124 | 0.065 | 0.210 |
| 02 | 0.181 | 0.087 | **0.050** | 0.096 | 0.086 | 0.161 | 0.109 | 0.231 |
| 03 | <u>0.195</u> | 0.094 | **0.024** | 0.057 | 0.078 | 0.125 | 0.107 | <u>0.321</u> |
| 04 | <u>0.163</u> | 0.096 | **0.038** | 0.078 | 0.061 | 0.126 | 0.106 | <u>0.333</u> |
| 05 | <u>0.227</u> | 0.075 | **0.028** | 0.055 | 0.041 | 0.099 | 0.090 | <u>0.386</u> |
| 06 | <u>0.237</u> | 0.068 | **0.031** | 0.064 | 0.059 | 0.099 | 0.082 | <u>0.360</u> |
| 07 | <u>0.129</u> | 0.065 | **0.021** | 0.050 | 0.061 | 0.130 | 0.093 | <u>0.450</u> |
| 08 | 0.045 | 0.031 | **0.022** | 0.048 | 0.052 | 0.109 | <u>0.112</u> | <u>0.581</u> |
| 09 | <u>0.412</u> | 0.079 | **0.021** | 0.048 | 0.043 | 0.076 | 0.050 | <u>0.271</u> |
| 10 | <u>0.226</u> | 0.055 | **0.014** | 0.047 | 0.034 | 0.111 | 0.070 | <u>0.442</u> |
| 11 | 0.104 | 0.035 | **0.013** | 0.033 | 0.043 | <u>0.152</u> | 0.124 | <u>0.497</u> |
| 12 | 0.113 | 0.065 | **0.039** | 0.089 | 0.092 | 0.131 | <u>0.152</u> | <u>0.319</u> |

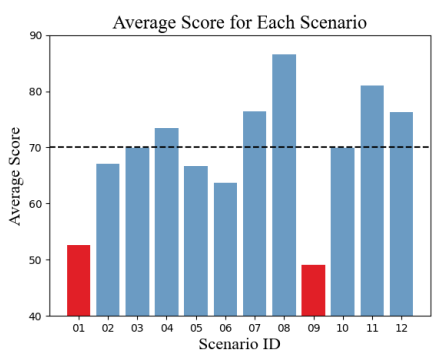

Figure 10. The average evaluation score for each scenario for simulator

Fig. 10 further illustrates the distribution of average scores across all simulator scenarios. It can be observed that over half of the scenarios have an average score below 70, reinforcing the aforementioned observation. Scenes 1 and 9, highlighted in red, exhibit the lowest average scores, aligning with their challenging setups. Scene 1 involves a sudden failure of the lead vehicle amid dense traffic, presenting a high level of challenge for the driver. Scene 9, although lacking apparent safety threats, involves interactions with various traffic participants while traversing an uncontrolled intersection, significantly influencing assessments.

## V. POTENTIAL APPLICATIONS

### A. Driver Intelligence Learning

We hope that a flexible combination of various types of data within the D2E dataset will unlock research opportunities, and this section will present potential applications of datasets as examples.

The drivers' eye-tracking data and physiological data can more intuitively and intrinsically reflect the drivers' response to challenging scenarios. For low-level autonomous driving, eye-tracking and physiological signals reflect the driver's state and further assess the need for takeover. For high-level autonomous driving, combining human factor data and driving environment data can explore the driver's behavioral mechanisms and help the decision-making algorithms to better learn human intelligence. For instance, the driver's concentration area can be used as training input to improve the learning model's ability to understand which regions are worthy of attention.

### B. Trajectory prediction, Planning and Control Learning

The D2E dataset provides more available input data except for traditional trajectory and control information, which benefits research about trajectory prediction, planning and control. The high-interaction driving events of the D2Edataset contain high-precision perceptual information, which can be directly used for training as true values. Meanwhile, human factor data provide additional training inputs. For instance, physiological signals can be used to determine the driver's intention of acceleration or deceleration, while eye-tracking signals can reflect the possible steering direction.

### C. Decision-making Evaluation Learning

Notably, the D2E dataset collects ratings for each driving event, providing interesting opportunities for research on decision evaluation. Utilizing subjective ratings and objective driving data makes it possible to calibrate the parameters of decision evaluation models such as weighting factors or develop novel models. Using the driver's physiological signals to judge when the driver is alertness and combining driving environment information, detailed research can also establish a risk perception model for decision-making or decision evaluation.

The above discussion serves as a starting point, more applications of the D2E dataset are expected to be explored by researchers.

## VI. CONCLUSION

In this paper, we propose the D2E, an autonomous decision-making dataset involving driver states and human evaluation. Collected from both driving simulators and real-world driving, D2E contains driver physiological signals, eye track attention distribution, first-person view videos, environment agents' information, and evaluation scores for each driving case given by third-party human raters. A high interaction level is achieved through pertinent scenario design according to driving risk mechanism and manual filtering. The distribution of drivers and raters covers a variety of features, ensuring comprehensiveness. Basic organization format, data processing, and analysis results are also given. The potential applications of D2E include the learning of driver intelligence and behavior, planning and control, and decision-making evaluation models.

The dataset will be released on GitHub and available for researchers to download and use once this paper is accepted.

## REFERENCES

[1] Y. Wang *et al.*, "Decision-Making Driven by Driver Intelligence and Environment Reasoning for High-Level Autonomous Vehicles: A Survey," *IEEE Transactions on Intelligent Transportation Systems*, 2023.

[2] S. Grigorescu, B. Trasnea, T. Cocias, and G. Macesanu, "A survey of deep learning techniques for autonomous driving," *Journal of Field Robotics*, vol. 37, no. 3, pp. 362-386, 2020.

[3] A. Sadat, M. Ren, A. Pokrovsky, Y.-C. Lin, E. Yumer, and R. Urtasun, "Jointly learnable behavior and trajectory planning for self-driving vehicles," in *2019 IEEE/RSJ International Conference on Intelligent Robots and Systems (IROS)*, 2019: IEEE, pp. 3949-3956.

[4] X. Liang, T. Wang, L. Yang, and E. Xing, "Cirl: Controllable imitative reinforcement learning for vision-based self-driving," in

*Proceedings of the European conference on computer vision (ECCV)*, 2018, pp. 584-599.

[5] Y. Wang, Z. Han, Y. Xing, S. Xu, and J. Wang, "A Survey on Datasets for the Decision Making of Autonomous Vehicles," *IEEE Intelligent Transportation Systems Magazine*, 2024.

[6] H. Caesar *et al.*, "NuPlan: A closed-loop ML-based planning benchmark for autonomous vehicles," Feb. 2022, doi: 10.48550/arXiv.2106.11810.

[7] F. Vicente, Z. Huang, X. Xiong, F. De la Torre, W. Zhang, and D. Levi, "Driver gaze tracking and eyes off the road detection system," *IEEE Transactions on Intelligent Transportation Systems*, vol. 16, no. 4, pp. 2014-2027, 2015.

[8] P. Sun *et al.*, "Scalability in perception for autonomous driving: Waymo open dataset," in *Proceedings of the IEEE/CVF conference on computer vision and pattern recognition*, 2020, pp. 2446-2454.

[9] W. Zhan *et al.*, "Interaction dataset: An international, adversarial and cooperative motion dataset in interactive driving scenarios with semantic maps," *arXiv preprint arXiv:1910.03088*, 2019.

[10] Y. Chen *et al.*, "DBNet: A Large-Scale Dataset for Driving Behavior Learning," *Retrieved August*, vol. 9, p. 2020, 2019.

[11] J. Binas, D. Neil, S.-C. Liu, and T. Delbruck, "DDD17: End-to-end DAVIS driving dataset," *arXiv preprint arXiv:1711.01458*, 2017.

[12] Z. Yan, L. Sun, T. Krajník, and Y. Ruichek, "EU Long-term Dataset with Multiple Sensors for Autonomous Driving," in *2020 IEEE/RSJ International Conference on Intelligent Robots and Systems (IROS)*, Oct. 2020, pp. 10697-10704, doi: 10.1109/IROS45743.2020.9341406.

[13] V. Punzo, M. T. Borzacchiello, and B. Ciuffo, "On the assessment of vehicle trajectory data accuracy and application to the Next Generation SIMulation (NGSIM) program data," *Transportation Research Part C: Emerging Technologies*, vol. 19, no. 6, pp. 1243-1262, Dec. 2011, doi: 10.1016/j.trc.2010.12.007.

[14] R. Krajewski, J. Bock, L. Kloeker, and L. Eckstein, "The highD Dataset: A Drone Dataset of Naturalistic Vehicle Trajectories on German Highways for Validation of Highly Automated Driving Systems," in *2018 21st International Conference on Intelligent Transportation Systems (ITSC)*, Nov. 2018, pp. 2118-2125, doi: 10.1109/ITSC.2018.8569552.

[15] J. F. Antin, S. Lee, M. A. Perez, T. A. Dingus, J. M. Hankey, and A. Brach, "Second strategic highway research program naturalistic driving study methods," *Safety Science*, vol. 119, pp. 2-10, Nov. 2019, doi: 10.1016/j.ssci.2019.01.016.

[16] A. Geiger, P. Lenz, and R. Urtasun, "Are we ready for autonomous driving? The KITTI vision benchmark suite," in *2012 IEEE Conference on Computer Vision and Pattern Recognition (CVPR)*, June 2012, Providence, RI: IEEE, pp. 3354-3361, doi: 10.1109/CVPR.2012.6248074.

[17] S. Xie, S. Chen, J. Zheng, M. Tomizuka, N. Zheng, and J. Wang, "From Human Driving to Automated Driving: What Do We Know About Drivers?," *IEEE Transactions on Intelligent Transportation Systems*, vol. 23, no. 7, pp. 6189-6205, July 2022, doi: 10.1109/TITS.2021.3084149.

[18] D.-w. Xing, X.-s. Li, X.-l. Zheng, Y.-y. Ren, and Y. Ishiwatari, "Study on driver's preview time based on field tests," in *2017 4th International Conference on Transportation Information and Safety (ICTIS)*, Aug. 2017, pp. 575-580, doi: 10.1109/ICTIS.2017.8047823.

[19] J. D. Ortega *et al.*, "DMD: A Large-Scale Multi-modal Driver Monitoring Dataset for Attention and Alertness Analysis," A. Bartoli and A. Fusiello, Eds., 2020, Cham: Springer International Publishing, in Lecture Notes in Computer Science, pp. 387-405, doi: 10.1007/978-3-030-66823-5_23.

[20] M.-F. Chang *et al.*, "Argoverse: 3d tracking and forecasting with rich maps," in *Proceedings of the IEEE/CVF conference on computer vision and pattern recognition*, 2019, pp. 8748-8757.

[21] R. Krajewski, T. Moers, J. Bock, L. Vater, and L. Eckstein, "The rounD Dataset: A Drone Dataset of Road User Trajectories at Roundabouts in Germany," in *2020 IEEE 23rd International Conference on Intelligent Transportation Systems (ITSC)*, Sept. 2020, pp. 1-6, doi: 10.1109/ITSC45102.2020.9294728.

[22] J. Bock, R. Krajewski, T. Moers, S. Runde, L. Vater, and L. Eckstein, "The inD Dataset: A Drone Dataset of Naturalistic Road User Trajectories at German Intersections," in *2020 IEEE Intelligent Vehicles Symposium (IV)*, Oct. 2020, pp. 1929-1934, doi: 10.1109/IV47402.2020.9304839.

[23] Z. Yang *et al.*, "SurfelGAN: Synthesizing Realistic Sensor Data for Autonomous Driving," in *2020 IEEE/CVF Conference on Computer Vision and Pattern Recognition (CVPR)*, June 2020, pp. 11115-11124, doi: 10.1109/CVPR42600.2020.01113.

[24] K. Wang, W. Zhang, Z. Feng, H. Yu, and C. Wang, "Reasonable driving speed limits based on recognition time in a dynamic low-visibility environment related to fog—A driving simulator study," *Accident Analysis & Prevention*, vol. 154, p. 106060, May 2021, doi: 10.1016/j.aap.2021.106060.

[25] W. Li *et al.*, "AADS: Augmented autonomous driving simulation using data-driven algorithms," *Science Robotics*, vol. 4, no. 28, p. eaaw0863, Mar. 2019, doi: 10.1126/scirobotics.aaw0863.

[26] Y. Wang *et al.*, "A Novel Integrated Decision-Making Evaluation Method Towards ICV Testing Involving Multiple Driving Experience Factors," in *2023 IEEE International Conference on Unmanned Systems (ICUS)*, Oct. 2023, pp. 1-7, doi: 10.1109/ICUS58632.2023.10318493.